\documentclass[11pt]{article}

\usepackage{times}
\usepackage{graphicx}
\usepackage{amsmath,amssymb}
\usepackage{amsthm}
\usepackage{booktabs}
\usepackage{multirow}
\usepackage{microtype}
\usepackage{hyperref}
\usepackage{caption}
\usepackage{subcaption}
\usepackage{natbib}
\usepackage[margin=1in]{geometry}

\theoremstyle{definition}

\title{Mitigating Topology Biases in Graph Diffusion via Counterfactual Intervention}

\author{
  Wendi Wang\thanks{This work was done during an internship at The Pennsylvania State University.} \\
  Purdue University \\
  \texttt{wang7341@purdue.edu}
  \and
  Jiaxi Yang \\
  Pennsylvania State University \\
  \texttt{jmy5701@psu.edu}
  \and
  Yongkang Du \\
  Pennsylvania State University \\
  \texttt{ybd5136@psu.edu}
  \and 
  Lu Lin \\
  Pennsylvania State University \\
  \texttt{lulin@psu.edu}
}

\date{}  

\begin{document}

\title{Mitigating topology biases in Graph Diffusion via Counterfactual Intervention}


\maketitle

\begin{abstract}
  Graph diffusion models have gained significant attention in graph generation tasks, but they often inherit and amplify topology biases from sensitive attributes (e.g. gender, age, region), leading to unfair synthetic graphs. Existing fair graph generation using diffusion models is limited to specific graph-based applications with complete labels or requires simultaneous updates for graph structure and node attributes, making them unsuitable for general usage. To relax these limitations by applying the debiasing method directly on graph topology, we propose \emph{\underline{Fair} \underline{G}raph \underline{\smash{Diff}}usion Model (\textit{FairGDiff})}, a counterfactual-based one-step solution that mitigates topology biases while balancing fairness and utility. In detail, we construct a causal model to capture the relationship between sensitive attributes, biased link formation, and the generated graph structure. By answering the counterfactual question ``Would the graph structure change if the sensitive attribute were different?'', we estimate an unbiased treatment and incorporate it into the diffusion process. \textit{FairGDiff} integrates counterfactual learning into both forward diffusion and backward denoising, ensuring that the generated graphs are independent of sensitive attributes while preserving structural integrity.
Extensive experiments on real-world datasets demonstrate that \textit{FairGDiff} achieves a superior trade-off between fairness and utility, outperforming existing fair graph generation methods while maintaining scalability. 
\end{abstract}

\section{Introduction}
In recent years, substantial efforts have been dedicated to generated graph generation across various domains, such as privacy-preserving graph synthesis \citep{fu2023privacy}, molecule generation \citep{jin2018junction}, traffic modeling \citep{yu2019real}, and code completion \citep{brockschmidt2018generative}. Among the diverse techniques developed for graph generation, graph diffusion models \citep{vignac2022digress, liu2023generative, kong2023autoregressive} has emerged as a leading technique for this task. These models are typically trained to reverse a diffusion process, which progressively corrupts training graphs by adding noise, enabling them to effectively model and reproduce the underlying distribution of training graphs. 

However, while diffusion models excel at capturing graph distributions, they are susceptible to inheriting and even amplifying \emph{topology biases} presented in the training graphs.
Topology biases frequently exist in real-world graphs due to the \emph{homophilic effect} caused by certain sensitive node attributes (e.g., gender, race, age) \citep{wang2022unbiased,dai2021say,ling2023learning}, where a pair of nodes is more likely to form an edge if they share the same sensitive node attribute value. Without proper intervention, the diffusion model inheriting biases can further lead to the generation of biased graphs, which when presented in critical tasks like fraud detection, can undermine trust and fairness in high-stake decision-making, posing severe ethical concerns. 



\begin{figure}[t]
    \centering
    {\includegraphics[width=0.45\textwidth]{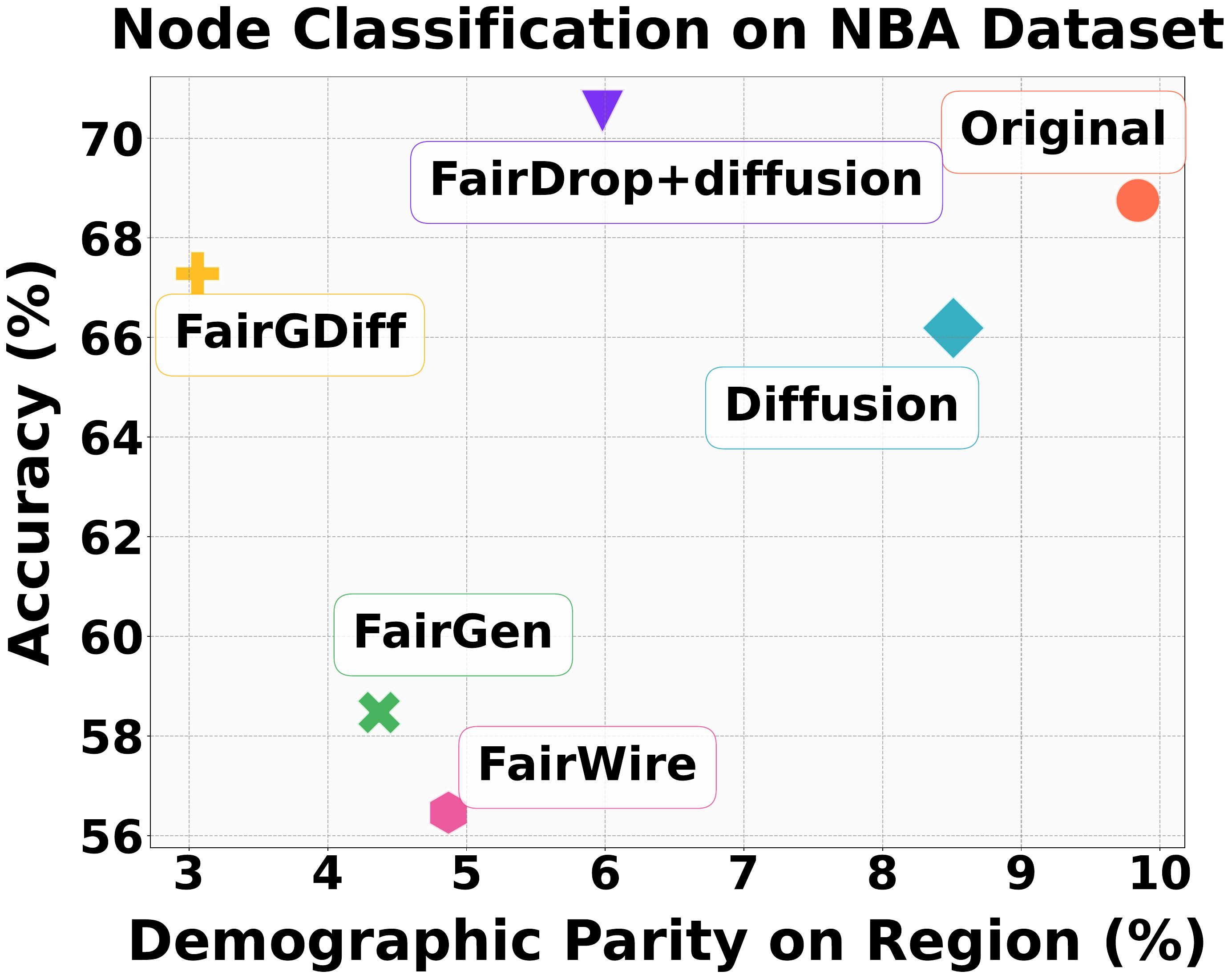}}
    {\includegraphics[width=0.45\textwidth]{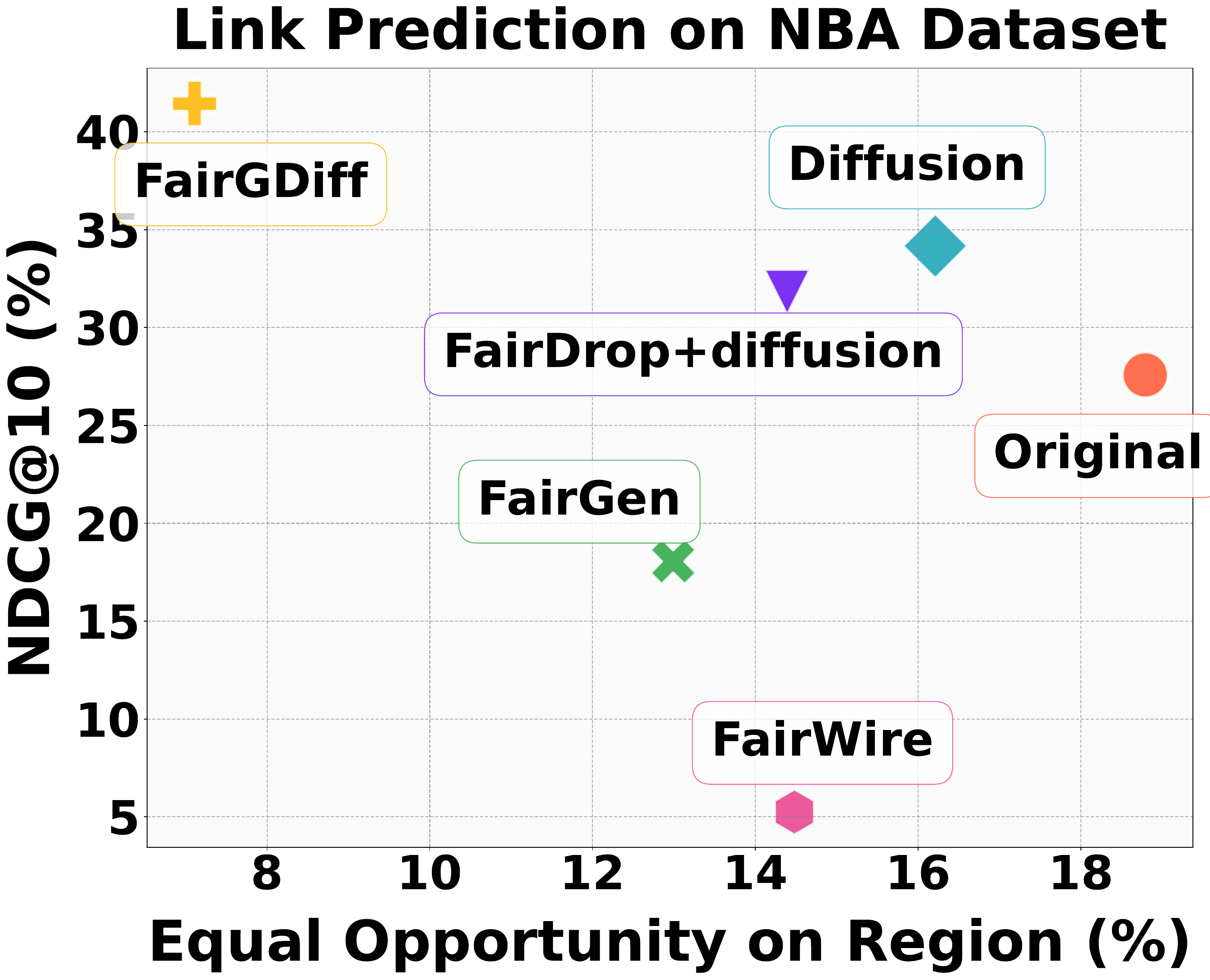}}
    \caption{
    Graphs generated by our FairGDiff achieve the best utility-fairness tradeoff in both supervised and unsupervised settings. \textbf{Left}: supervised node classification task. \textbf{Right}: unsupervised contrastive learning evaluated by downstream link prediction task. }
    \label{fig:tradeoff}
\end{figure}

Mitigating topology biases in graph generation is a crucial yet largely underexplored challenge. One straightforward solution is to utilize a \textbf{two-step} strategy, which perform graph structure purification (such as FairDrop~\citep{spinelli2021fairdrop}) before model training. However, this approach is often found to be suboptimal, as the quality of the generated graphs heavily relies on the purification method. Additionally, modifying the original training distribution through such preprocessing can degrade the utility of both the diffusion model and the generated graphs. This issue can be further evidenced in Figure~\ref{fig:tradeoff}, in which we compare the downstream performance of original graphs, generated graphs by vanilla diffusion trained on the original graphs (i.e., Original + Diffusion) and generated graphs via this two-step solution (i.e., FairDrop + Diffusion). The results demonstrate that such two-step strategies often fail to generate graphs that achieve a balanced trade-off between utility and unbiasedness for downstream usage.

Motivated by the drawbacks of two-step strategies for fair graph generation mentioned above, some recent works, such as FairWire~\citep{kose2024fairwire} and FairGen~\citep{zheng2024fairgen}, attempt to enforce fairness during the training of graph generative models, rather than relying on the pre-processing step in two-step approaches.
However, FairWire~\citep{kose2024fairwire} requires simultaneous updates for graph structure, node features, and sensitive attribute information to ensure bias mitigation, which may limit its applications when the generated features are substantially different from node features in downstream tasks. FairGen~\citep{zheng2024fairgen} requires label information to perform the debiasing method via random walk, which is also a limitation for general graphs where labels are unavailable or incomplete. These limitations highlight the need for fairness-aware graph generation methods that operate effectively without requiring explicit attribute modifications or label dependencies.

In this work, we aim to relax these limitations by directly disentangling the sensitive attribute based homophilic pattern from the graph topology when generating the synthetic graph, which results in a \textbf{general} \textbf{one-step} solution. In other words, it is not necessary to acquire an unbiased input graph using graph purification methods before diffusion model training, and such disentanglement and diffusion model training occur simultaneously. By saying it general, since the homophilic pattern only concerns biased edges between nodes with similar sensitive attributes, its disentanglement neither requires label information nor interferes the original node features. Specifically, we disentangle the homophilic pattern on sensitive attributes through a counterfactual solution, such that a \emph{\underline{Fair} \underline{G}raph \underline{\smash{Diff}}usion Model (\textit{FairGDiff})} is achieved for unbiased graph generation. 
From causal inference perspectives, \textit{FairGDiff} integrates causal modeling into graph diffusion to mitigate the bias introduced by sensitive attributes. In the graph generative process, we construct a causal model and a sensitive attribute treatment to estimate the effect of topology biases on the edge probability between two nodes. 
By answering a counterfactual question ``would the edge change if the sensitive attribute of corresponding nodes were different?'', we control the sensitive attribute treatment to generate interventional guidance for  the diffusion process. \textit{FairGDiff} then utilizes counterfactual-guided training to ensure that the generated graphs remain independent of sensitive attributes while maintaining key structural properties. Therefore, it achieves better trade-off between downstream utility and fairness, as shown in Figure \ref{fig:tradeoff}. 

The main \textbf{C}ontributions of our paper are listed below: \\
\noindent\textbf{C1:} We propose \textit{FairGDiff}, a general one-step counterfactual solution to mitigate topology biases in graph diffusion while balancing fairness and utility. From a causal inference perspective, we construct a causal model to capture the interplay between sensitive attributes, biased link formation, and the generated graph, enabling counterfactual reasoning for unbiased treatment estimation. \\
\noindent\textbf{C2:} We integrate counterfactual treatments into the graph diffusion process, where both factual and counterfactual graphs undergo noise injection and denoising, ensuring that the generated graphs remain independent of sensitive attributes while preserving structural integrity. \\
\noindent\textbf{C3:} Extensive experiments demonstrate that our approach achieves an optimal trade-off between fairness and utility compared to existing methods while maintaining high efficiency and scalability.

\section{Related Work}
Graph fairness learning has garnered significant attention in recent years~\citep{chen2024fairness,laclau2022survey}. Current efforts in graph fairness aim to address three key notions: group fairness~\citep{dai2021say,ijcai2019p456}, individual fairness~\citep{song2022guide}, and counterfactual fairness~\citep{guo2023towards}. To mitigate bias in graph learning, existing methods typically intervene at one of three stages: (1) pre-processing the graph before training~\citep{li2021dyadic,chen2022graph}, (2) modifying the training procedure itself~\citep{li2021dyadic}, or (3) applying post-processing techniques to adjust the outputs~\citep{masrour2020bursting}.
Although promising progress has been made in graph fairness learning, most existing studies focus on node classification, link prediction, and graph classification tasks. This leaves a significant research gap in addressing fairness within graph generative models.

A limited study, FairDrop~\citep{li2021dyadic}, explores graph fairness by deleting suspicious biased edges of a certain probability. Yet, this approach is often suboptimal, as randomly deleting some edges can distort the training distribution and reduce the quality of generated graphs. To address these limitations, FairWire~\citep{kose2024fairwire} jointly updates graph structure, node features, and sensitive attributes to mitigate bias. However, this may restrict its applicability when the generated features differ significantly from those required in downstream tasks. 
Another study, FairGen~\citep{zheng2024fairgen}, relies on label information to apply its debiasing method via random walks, presenting a limitation in scenarios where labels are unavailable or incomplete. 
These limitations motivate our proposed fairness-aware graph generation method that avoids reliance on attribute changes or label information.

\section{Notations and Preliminaries}
\label{sec:method1}
\subsection{Notations}
Let $G=(\mathcal{U}, A)$ be an undirected attributed graph and $\mathcal{V}=\{ v_{i} \}^{n}_{i=1}$ be the set of $n$ nodes constructing this graph. $A \in {\mathrm{R}}^{n \times n}$ denotes the adjacency matrix. We define $\mathcal{U}=\{ {\mathbf{u}}_{i} \}^{n}_{i=1}$, where ${\mathbf{u}}_i$ is a $d$-dimensional attribute value vector for each node $v_i \in \mathcal{V}$. We assume all attributes are categorical and binary with the first attribute as the sensitive attribute and the rest $d-1$ attributes as non-sensitive attribute. 
We denote the node feature matrix $X \in {\mathrm{R}}^{n \times {(d-1)}}$ as a set of all non-sensitive attribute vectors, i.e. $X = \{{\mathbf{u}}_i[1 :]\}^{n}_{i=1}$. The sensitive attribute values of all nodes in $\mathcal{V}$ consist of a sensitive attribute vector $\textbf{s} = \{{\mathbf{u}}_{i}[: 1]\}^{n}_{i=1}$, where ${\mathbf{u}}_i[: 1] \in \{0, 1\}$ for each node $v_i \in \mathcal{V}$. 

\subsection{Graph Diffusion Model}\label{subsec:graph_diffusion}
Graph diffusion models have emerged as a class of widely adopted graph generative methods across various domains, which work by iteratively refining the graph information (e.g. structure, attribute or representation) to simulate a stochastic diffusion process. There are four main categories of graph diffusion models, namely denoising diffusion probabilistic models (DDPMs) \citep{jin2018junction, bojchevski2018netgan, vignac2022digress, haefeli2022diffusion, anand2022protein, trippe2022diffusion, luo2022antigen}, score-based generative models (SGMs) \citep{niu2020permutation, chen2022nvdiff}, latent diffusion models \citep{yang2024graphusion, evdaimon2024neural, cai2024latent}, and stochastic differential equation based diffusion models (SDEs) \citep{luo2023fast, jo2022score}, differing in the design of the \textbf{forward diffusion process} and \textbf{backward denoising process}, which in general can be summarized as follows. 

\paragraph{\textbf{Forward Diffusion Process (Noise Injection).}}
The forward process gradually corrupts a given graph by adding noise over multiple timesteps. 
Given an initial graph $G_0$, the process iteratively adds Gaussian noise to transform it into a nearly random graph. We use $z_0$ to denote the initial graph information, which can either be the original graph or a latent encoding of the graph. For graph diffusion models on graph domain like DDPMs~\citep{jin2018junction, bojchevski2018netgan, vignac2022digress, haefeli2022diffusion, anand2022protein, trippe2022diffusion, luo2022antigen}, SGMs~\citep{niu2020permutation, chen2022nvdiff}, and SDEs~\citep{luo2023fast, jo2022score}, the noise is added directly to the input graph i.e., $z_0\colon = G_0=(\mathbf{X}_0, \mathbf{A}_0)$; and for latent graph diffusion models \citep{yang2024graphusion, evdaimon2024neural, cai2024latent}, the noises are injected to the latent space where the input graph is embedded via an autoencoder, i.e., $z_0 \colon = E(G_0)$. Taking in the initial graph information $z_0$, 
at each time step $t$, the diffusion process follows: 
\begin{equation}
    q({z}_t|{z}_{t-1})=\mathcal{N}({z}_t; \sqrt{1-\beta_{t}} {z}_{t-1}, \beta_{t} I),
    \label{eq:forward}
\end{equation}
where $\beta_{t}$ is a known variance schedule with $0<\beta_{t}<1$, $I$ represents the identity matrix or the unit vector according to the category of graph diffusion models. Finally, the noisy graph information $z_t$ is obtained as follows: 
\begin{equation}
    q({z}_t|{z})=\mathcal{N}({z}_t; \sqrt{{\alpha}'_{t}} {z}, (1-{\alpha}'_{t}) I),
    \label{eq:forward_t}
\end{equation}
where ${\alpha}'_t=\prod_{i=1}^{t} \alpha_i, \alpha_t=1-\beta_t$.

\paragraph{\textbf{Backward Denoising Process (Graph Generation).}}
The goal of the backward process is to reverse the noise and reconstruct a valid graph from the near-random one $z_T$. To this end, a denoising neural network $\epsilon_{\theta}$ is trained to predict the noise $\epsilon$ added at each time step $t$ by minimizing the objective:
\begin{align}
    {\mathcal{L}}_{\text{diffusion}} &= \mathbb{E}_{z_0, \epsilon \sim \mathcal{N}(0,1), t} \left[ || \epsilon -  \epsilon_{\theta} ({z}_t,t) ||_2^2 \right],
\end{align} 
which aim to minimize the difference between the actual noise and the predicted noise. By learning the noise distribution, the model can generated graphs by iteratively denoising from pure noise.
If conditioned on local or global properties of the output synthetic graph, such as node number, edge number, and global clustering coefficients, the graph diffusion model can be extended to learn \textbf{conditional distributions} as follows:
\begin{equation}
    \begin{split}
    {\mathcal{L}}_{\text{conditional}} &= \mathbb{E}_{z_0, \epsilon \sim \mathcal{N}(0,1), t} \left[ || \epsilon -  \epsilon_{\theta} ({z}_t,t,{\tau}_{\theta}(c)) ||_2^2 \right], 
    \label{eq:cond}
\end{split}
\end{equation} 
where $c$ denotes a property of $G$, and ${\tau}_{\theta}$ is a neural network jointly trained with ${\epsilon}_{\theta}$ to generate generated graphs  with property $c$.



\subsection{Causal Model}
\begin{figure}
    \centering
    \includegraphics[width=0.65\linewidth]{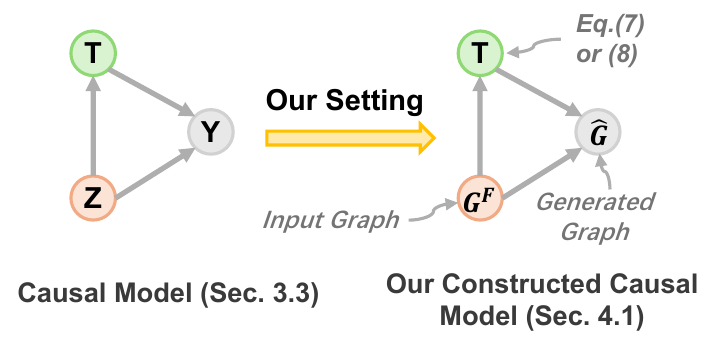}
    \caption{Construction of Causal Model for \textit{FairGDiff}.}
    \label{fig:causal_model}
\end{figure}

Counterfactual causal inference aims to find the causal relationship between treatment and outcome. A key concept in causal inference is counterfactual reasoning, which asks, ``Would the outcome have been different if the treatment had changed''. A fundamental goal of causal inference is to estimate the causal effect of a treatment T on an outcome Y. Given a confounder $Z$ that influences both $T$ and $Y$, a common causal structure is represented as:
$Z \rightarrow T \rightarrow Y$, as show in Figure~\ref{fig:causal_model}.

\section{Diffusion-based Fair Graph Generation}
\label{sec:method}
\paragraph{\textbf{Overview.}}
\begin{figure*}
    \centering
    \includegraphics[width=0.99\linewidth]{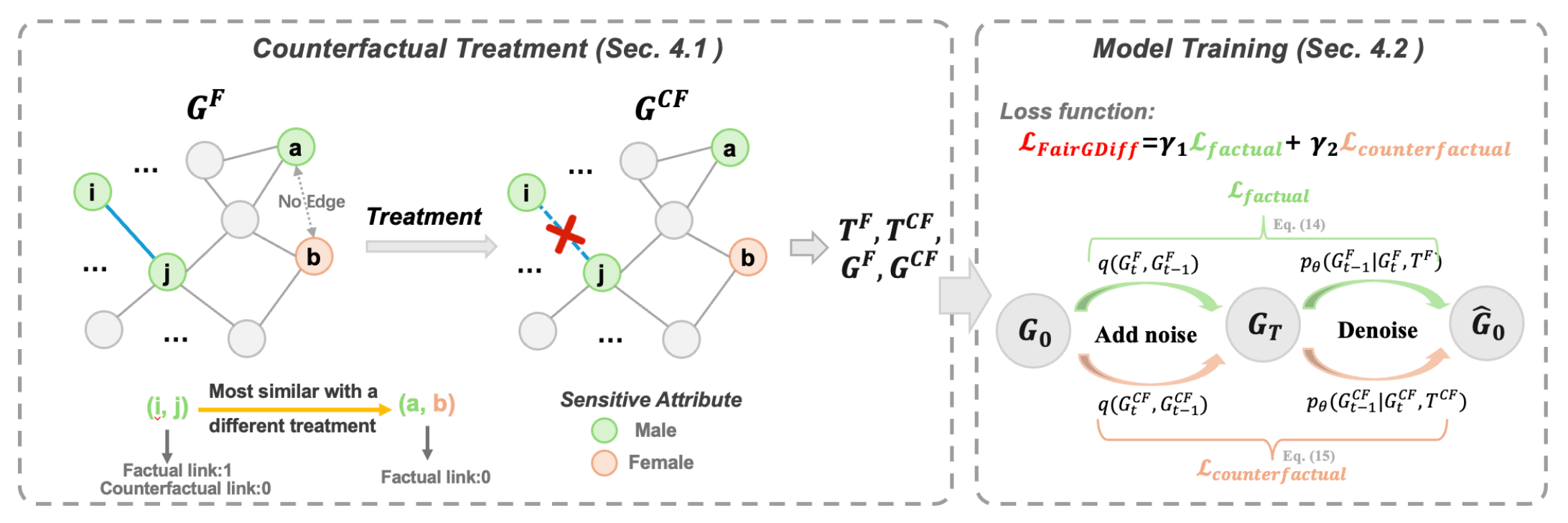}
    \caption{Overview of the proposed \textit{FairGDiff}, which consists two steps of Constructing Counterfactual Treatment and \textit{FairGDiff} training, illustrated with gender as the sensitive attribute. }
    \label{fig:overview}
\end{figure*}
This section first derives our definition of topology biases and targeted fairness, followed by introducing our proposed fairness solution for graph diffusion, termed \textbf{F}air \textbf{G}raph \textbf{Diff}usion (\textit{FairGDiff}). Figure~\ref{fig:overview} demonstrates the overview of \textit{FairGDiff}, which primarily comprised two modules: unbiased counterfactual treatment and bias-mitigated graph diffusion training.
In detail, inspired by causal inference, we estimate an unbiased counterfactual treatment by constructing a causal model that removes the influence of sensitive attributes on link formation. Then, we incorporate this treatment into the graph diffusion training process, ensuring that the generated graphs are structurally fair while preserving key properties of the original graph.
Finally, we analyze the computational complexity of \textit{FairGDiff}, demonstrating its efficiency in large-scale graph generation.

\subsection{Bias Analysis and Targeted Fairness}
This section derives our definition of topology biases and targeted fairness. Among diverse kinds of topology bias (sensitive-attribute homophily, degree assortativity, preferential attachment bias, etc.), we focus on \textit{sensitive attribute homophily} (nodes with same sensitive attributes are more likely to form an edge in between), a dominant type of topology biases in graph learning and social networks. We quantify topology biases of a graph $G$ (either the input or the synthetic graph) with the ratios 
\begin{equation}
    \mathcal{T}_0(G)=\frac{\mathbb{P}(\hat{A}_{ij}=0|s_i=s_j)}{\mathbb{P}(\hat{A}_{ij}=0|s_i \neq s_j)}, \mathcal{T}_1(G)=\frac{\mathbb{P}(\hat{A}_{ij}=1|s_i=s_j)}{\mathbb{P}(\hat{A}_{ij}=1|s_i \neq s_j)}.
\end{equation}

Given a biased input graph $G^{F}=(\mathcal{U}, A^{F})$, our goal is to generate a fair graph structure $\hat{G}=(\hat{\mathcal{U}}, \hat{A})$ independent of sensitive attributes, thereby mitigating topology biases arising from sensitive attribute homophily. To formalize this, a topology-fair graph structure $\hat{G}=(\hat{\mathcal{U}}, \hat{A})$ can be defined as:
\begin{equation}
    \mathcal{T}_x(\hat{G})=\frac{\mathbb{P}(\hat{A}_{ij}=x|s_i=s_j)}{\mathbb{P}(\hat{A}_{ij}=x|s_i \neq s_j)}=\frac{\frac{\text{Freq}(\hat{A}_{ij}=x, s_i=s_j)}{\text{Freq}(\hat{A}_{ij}=x, s_i \neq s_j)}}{\frac{\text{Freq}(s_i=s_j)}{\text{Freq}(s_i \neq s_j)}}=1,x\in\{0,1\},
    \label{eq:topo_bias}
\end{equation}
which indicates that $\hat{G}$ is independent of sensitive attributes. 

During the graph generation, we only modify the sensitive attributes to be counterfactual, while all non-sensitive attribute values are preserved:
\begin{equation}
    \mathcal{U}[1:]=\hat{\mathcal{U}}[1:],
\end{equation}
The above preservation of non-sensitive attributes is considered to be a novelty of our method compared to previous works. 

\subsection{Unbiased Counterfactual Treatment with Causal Model}
\label{subsec:counterfactual}
\paragraph{\textbf{Construction of Causal Model.}}
To mitigate the effect of topological bias in graph diffusion, we construct a causal model as shown in Figure~\ref{fig:causal_model} that captures the dependency between biased input graph $G^F=(\mathcal{U},A^F)$, treatment of the link between two nodes on the graph, and the generated graph $\hat{G}=(\hat{\mathcal{U}},\hat{A})$. In our causal setting, the confounder $Z$ is the biased input graph $G^F$, which influences both the treatment assignment and the structure of the generated graph. 
The treatment variable $T$ determines whether the link formation process is biased or not. In a topology-biased setting, edges preferentially form between nodes with the same sensitive attribute. Conversely, an unbiased link formation process is independent of sensitive attribute homophily.
And the outcome $Y$ in the causal model is the generated graph $\hat{G}$ by the trained graph diffusion model, which inherits the biased or unbiased property from $T$. This causal dependency follows the structure of causal model $Z \rightarrow T \rightarrow Y$, where $Z$ influences $T$ by increasing or decreasing the likelihood of homophilic connections between node pairs on the graph, which in turn affects outcome (i.e. generated graph $\hat{G}$) by reinforcing structural biases in the generated graph. 
Therefore, in order to achieve the goal of training a fair graph diffusion model, we attempt to address the following key counterfactual question:
\emph{``Given a biased input graph  $G^F = (\mathcal{U}, A^F)$  and the identified bias  $T^F$ , would the graph structure  $A^F$  change if the sensitive attribute homophily (i.e. $T^F$ ) were different?''}

\paragraph{\textbf{Treatment Variable.}}
Numerous studies have shown that nodes sharing the same sensitive attribute are more likely to form edges, reinforcing homophily-induced bias~\citep{zhao2022learning,zheng2024fairgen,kose2024fairwire}. Such treatment assignments fail to capture a fair structural representation, as they entangle sensitive attributes with the underlying graph topology. To formalize this, we define the binary treatment variable $T$, which determines if a node pair $(v_i, v_j)$ forms an edge based on their sensitive attributes and the factual treatment matrix $T^F$ is defined as follows:
\begin{equation}
    T^{F}_{ij} =
    \begin{cases} 
    1, & \text{if } \mathbf{s}_i=\mathbf{s}_j \text{ for } v_i, v_j \in \mathcal{V},\\
    0, & \text{otherwise. }
    \end{cases}
\label{eq:T_f}
\end{equation}

To mitigate this bias, we construct a counterfactual treatment assignment  $T^{CF}$ by finding an alternative treatment independent of the sensitive attribute. Instead of simply inverting  $T^F$, we estimate $T^{CF}$ by finding the most similar unbiased node pair in the graph. Specifically, for each node pair $(v_i, v_j)$ with a biased indicator $T^F_{ij}$, we find another nearest node pair $(v_a, v_b)$ with the opposite treatment assignment  $T^F_{ab}$. We use Euclidean distance function $d$ to measure the distance among node pairs. If such a bias-free context exists within a certain similarity threshold $\xi$, we use $T^F_{ab}$ to estimate the unbiased indicator $T^{CF}_{ij}$:
\begin{equation}
    \begin{aligned}
        T^{CF}_{ij} &= T^F_{ab},\\
    \end{aligned}
\end{equation}
where $ (v_a, v_b) = \arg\min_{(v_p, v_q)} d((v_i, v_j), (v_p, v_q))$, and $d((v_i, v_j)$, $(v_p, v_q)) \leq \xi$.
This ensures that $T^{CF}$ preserves the structural similarities among non-sensitive node attributes while eliminating bias due to sensitive attributes.

\paragraph{\textbf{Counterfactual Treatment.}}
Once the unbiased counterfactual treatment $T^{CF}$ is estimated, we can calculate the counterfactual adjacency matrix $A^{CF}$ using $T^{CF}$ to represent a debiased graph structure. The counterfactual structure is estimated using a learned function $f$ that models link formation as:
\begin{equation} 
    A^{CF} = f(T^{CF}, X),
\end{equation}

where $X$ represents node attributes excluding sensitive features. The most similar observed context algorithm ensures that $A^{CF}$ is independent of the sensitive attribute, as it only considers non-sensitive node similarities when estimating new links. Since both $T^{CF}$ and $A^{CF}$ are derived based on bias-free node contexts, the counterfactual graph structure achieves unbiasedness while maintaining the fundamental topology of the original graph.

By integrating both factual and counterfactual structures into model training, we effectively remove bias in graph diffusion, ensuring that node representations are not skewed by sensitive attributes. The details of the counterfactual estimation process are summarized in Algorithm 1 in the appendix.

\subsection{FairGDiff Training with Counterfactual Treatment}
\label{subsec:model_training}
To generate fair graph structures while preserving utility, \textit{FairGDiff} integrates counterfactual treatment throughout the entire diffusion process, ensuring that both forward diffusion and backward denoising incorporate counterfactual learning. In the forward diffusion phase, noise is progressively added to both the factual and counterfactual graph representations, ensuring that the model learns structural variations under different treatment assignments. During the backward denoising phase, the model reconstructs the graph while conditioning on both factual and counterfactual treatments, effectively mitigating bias inherited from sensitive attributes. By leveraging both factual and counterfactual diffusion paths, \textit{FairGDiff} ensures that the generated graph structure remains independent of sensitive attributes while preserving key topological properties.

\paragraph{\textbf{Forward Diffusion of FairGDiff}} 
The forward diffusion process follows a Markovian structure, progressively adding noise to the graph over multiple timesteps. Specifically, given the factual graph representation $G^F$ and the counterfactual graph representation $G^{CF}$, the diffusion process of \textit{FairGDiff} is defined as follows:
\begin{equation}
    q({z}_t^{F}|{z}_{t-1}^{F})=\mathcal{N}({z}_t^{F}; \sqrt{1-\beta_{t}} {z}_{t-1}^{F}, \beta_{t} I),
    \label{eq:forward_f}
\end{equation}
\begin{equation}
    q({z}_t^{CF}|{z}_{t-1}^{CF})=\mathcal{N}({z}_t^{CF}; \sqrt{1-\beta_{t}} {z}_{t-1}^{CF}, \beta_{t} I),
    \label{eq:forward_cf}
\end{equation}
where $\beta_{t}$ is a known variance schedule with $0<\beta_{t}<1$. Finally, the noisy graph information $z_t^{F}$ and $z_t^{CF}$ are further obtained as: 
\begin{equation}
    q({z}_t^{F}|{z}^{F})=\mathcal{N}({z}_t^{F}; \sqrt{\Bar{\alpha}_{t}} {z}^{F}, ({1-\Bar{\alpha}}_{t}) I),
    \label{eq:forward_t_f}
\end{equation}
\begin{equation}
    q({z}_t^{CF}|{z}^{CF})=\mathcal{N}({z}_t^{CF}; \sqrt{\Bar{\alpha}_{t}} {z}^{CF}, ({1-\Bar{\alpha}}_{t}) I),
    \label{eq:forward_t_cf}
\end{equation}
where $\Bar{\alpha}_t=\prod_{i=1}^{t} \alpha_i, \alpha_t=1-\beta_t$.

During training the denoising model, we jointly train $\epsilon_{\theta}$ and a treatment encoder ${\tau}_{\theta}$ by setting ${c}^{F} = \tau_{\theta} (T^{F}, X)$ and ${c}^{CF} = \tau_{\theta} (T^{CF}, X)$. Again, because node feature matrix $X$ is independent of the sensitive attribute, ${c}^{F}$ inherits the bias from input graph structure through $T^{F}$ and ${c}^{CF}$ is an unbiased vector combining information in $T^{CF}$ and $X$. 

\paragraph{\textbf{Backward Denoising of FairGDiff}}
The backward process reconstructs the graph by learning to denoise while incorporating both factual and counterfactual treatments. To achieve this, FairGDiff trains a denoising network $\epsilon_\theta$ along with a treatment encoder $\tau_\theta$ to condition the process on treatment assignments:
\begin{equation}
    \mathcal{L}_{\text{factual}} =  \mathbb{E}_{E(G^{F}), \epsilon \sim \mathcal{N}(0,1), t} [|| \epsilon -  \epsilon_{\theta} ({z}_t^{F},t,{c}^{F}) ||_2^2],
\end{equation} 
\begin{equation}
    \mathcal{L}_{\text{counterfactual}} = \mathbb{E}_{E(G^{CF}), \epsilon \sim \mathcal{N}(0,1), t} [|| \epsilon - \epsilon_{\theta} ({z}_t^{CF},t,{c}^{CF}) ||_2^2].
\end{equation} 
Here, $c^F$ and $c^{CF}$ represent the conditioning signals derived from the factual and counterfactual treatments, respectively, ensuring that the denoised graph structure aligns with bias-free representations.

\paragraph{\textbf{Training Objective of FairGDiff}} \textit{FairGDiff} obtains an unbiased output graph structure $\hat{A}$ via optimizing two objectives: (1 maintaining high utility in $\hat{A}$; (2 intervening bias inheritance in $\hat{A}$. Specifically, we combine $\mathcal{L}_{\text{factual}}$ with $\mathcal{L}_{\text{counterfactual}}$ to maintain high utility in $\hat{A}$ when performing debiasing through ${c}^{CF}$ intervention. Therefore, the overall objective of \textit{FairGDiff} aims to optimize the following function:
\begin{equation}
    \mathcal{L} = \gamma_1 \mathcal{L}_{\text{factual}} + \gamma_2 \mathcal{L}_{\text{counterfactual}},
\end{equation} 
with $\gamma_1$ and $\gamma_2$ as hyperparameters for maintaining graph utility and mitigating bias inheritance. 


\subsection{Complexity Analysis} \label{sec:method3}


The complexity of the Variational Graph Autoencoder~\citep{evdaimon2024neural} is determined by a GIN encoder and a three-layer MLP decoder, with a complexity of $\mathcal{O}(Ld^2N + Ld|\mathcal{E}|)$ \citep{wu2020comprehensive} and $\mathcal{O}(d_h (d+2) |\mathcal{E}|)$ \citep{zhao2022learning} respectively, where $L$ is the number of GIN layers, $d$ is the number of attributes, $|\mathcal{E}|$ is the number of edges, and $d_h$ is the number of neurons in the hidden layer. The complexity of the Latent Diffusion Model is limited to $\mathcal{O}(T l^2)$ where $l$ is the dimension of the latent representation $z$ and $T$ is the denoising time step. Therefore, the proposed Unbiased Graph Generator has linear time complexity w.r.t. the sum of node and edge counts.







\begin{table*}[ht]
\centering
\resizebox{\textwidth}{!}{%
\begin{tabular}{ccccccccccc}
\toprule
\textbf{Dataset} & \textbf{Metric (\%)} & \textbf{MLP} & \textbf{GCN} & \textbf{FairAdj} & \textbf{FairDrop (0.5)} & \textbf{FairDrop (0.75)} & \textbf{FairGen} & \textbf{FairWire} & \textbf{Diffusion} & \textbf{FairGDiff} \\ 
\midrule
\multirow{4}{*}{NBA} 
& Accuracy & 65.35 ± 1.11 & \textbf{72.06 ± 1.48} & 66.18 ± 0.65 & 70.57 ± 1.68 & \textbf{70.75 ± 1.79} & 70.52 ± 1.51 & 62.56 ± 0.23 & 68.30 ± 0.33 & 68.08 ± 0.00 \\ 
& $\Delta_{DP}$ & 7.53 ± 2.76 & 7.24 ± 6.01 & 12.22 ± 1.54 & 8.67 ± 5.04 & 7.70 ± 3.46 & 5.88 ± 2.78 & \textbf{2.06 ± 1.17} & 6.68 ± 1.68 & \textbf{1.79 ± 1.00} \\ 
& $\Delta_{EO}$ & 5.76 ± 2.62 & 10.55 ± 5.02 & \textbf{2.45 ± 1.61} & 20.40 ± 5.53 & \textbf{3.54 ± 2.44} & 10.62 ± 4.51 & 11.01 ± 2.57 & 14.57 ± 2.62 & 4.84 ± 1.74 \\ 
\midrule
\multirow{4}{*}{Pokec-n} 
& Accuracy & 62.77 ± 0.29 & \textbf{67.10 ± 0.92} & \textbf{66.84 ± 0.52} & 62.78 ± 1.27 & 59.68 ± 0.98 & 55.18 ± 0.24 & $-^{\ast}$ & 66.77 ± 0.69 & 65.91 ± 0.45 \\ 
& $\Delta_{DP}$ & 3.04 ± 0.48 & 2.83 ± 2.34 & 3.00 ± 2.40 & 2.17 ± 1.48 & 4.31 ± 2.20 & \textbf{0.81 ± 0.43} & $-^{\ast}$ & 2.31 ± 1.22 & 1\textbf{.94 ± 0.87} \\ 
& $\Delta_{EO}$ & 3.23 ± 0.41 & 2.82 ± 1.80 & 2.03 ± 1.44 & 5.80 ± 1.65 & 5.29 ± 2.32 & \textbf{0.72 ± 0.54} & $-^{\ast}$ & 1.66 ± 1.06 & \textbf{1.28 ± 0.81} \\ 
\midrule
\multirow{4}{*}{Pokec-z} 
& Accuracy & 64.47 ± 0.49 & \textbf{67.80 ± 1.43} & \textbf{67.87 ± 1.15} & 62.92 ± 0.94 & 62.76 ± 1.03 & 67.89 ± 0.59 & $-^{\ast}$ & 66.73 ± 0.50 & 66.80 ± 0.49 \\ 
& $\Delta_{DP}$ & 2.62 ± 0.85 & 3.74 ± 1.17 & 3.12 ± 0.94 & 4.45 ± 2.02 & 10.80 ± 1.59 & \textbf{2.03 ± 1.02} & $-^{\ast}$ & 2.47 ± 1.49 & \textbf{1.57 ± 0.99} \\ 
& $\Delta_{EO}$ & 4.61 ± 0.74 & 2.87 ± 1.62 & 2.36 ± 1.41 & 5.01 ± 2.29 & 10.01 ± 1.55 & \textbf{1.18 ± 0.68} & $-^{\ast}$ & 2.95 ± 1.55 & \textbf{1.11 ± 0.76} \\ 
\bottomrule 
\end{tabular}
}
\caption{Performance Comparisons of Node Classification on Synthetic Graphs. Results with the top two values of all metrics among all methods are highlighted in bold. *FairWire~\citep{kose2024fairwire} uses smaller-scaled Pokec-n and Pokec-z datasets compared to our FairGDiff. }
\label{tab:additional_results}
\end{table*}

\begin{table*}[ht]
\centering
\resizebox{\textwidth}{!}{%
\begin{tabular}{cccccccccc}
\toprule
\textbf{Dataset} & \textbf{Metric (\%)} & \textbf{GCL} & \textbf{Graphair} & \textbf{FairAdj} & \textbf{FairDrop (0.5)} & \textbf{FairDrop (0.75)} & \textbf{FairGen} & \textbf{Diffusion} & \textbf{FairGDiff} \\ 
\midrule
\multirow{3}{*}{NBA} & Accuracy & 70.78 ± 1.92 & \textbf{72.06 ± 1.46} & 57.16 ± 2.52 & \textbf{68.65 ± 0.69} & 67.23 ± 1.38 & 55.46 ± 2.59 & 64.82 ± 1.53 & 59.43 ± 5.06 \\ 
& $\Delta_{DP}$ & 9.60 ± 2.17 & 22.70 ± 1.93 & 16.57 ± 4.62 & \textbf{9.27 ± 3.99} & 14.09 ± 2.51 & 9.30 ± 4.74 & 22.84 ± 3.45 & \textbf{2.87 ± 3.71} \\ 
& $\Delta_{EO}$ & 32.27 ± 1.62 & 40.57 ± 4.15 & 27.44 ± 12.27 & 23.90 ± 3.65 & \textbf{3.79 ± 3.96} & 17.12 ± 7.21 & 35.74 ± 6.33 & \textbf{2.83 ± 1.89} \\ 
\midrule
\multirow{3}{*}{Pokec-z} & Accuracy & 61.97 ± 0.49 & \textbf{62.52 ± 0.34} & \textbf{67.37 ± 0.14} & 56.86 ± 0.78 & 56.77 ± 0.64 & 52.70 ± 1.51 & 59.07 ± 1.47 & 59.22 ± 0.64 \\ 
& $\Delta_{DP}$ & 2.37 ± 0.91 & 5.88 ± 1.10 & 1.34 ± 0.29 & 2.25 ± 0.70 & 1.44 ± 1.03 & \textbf{0.63 ± 0.35} & 2.88 ± 1.20 & \textbf{1.16 ± 0.48} \\ 
& $\Delta_{EO}$ & 3.44 ± 0.74 & 7.87 ± 1.12 & 3.13 ± 0.55 & 1.90 ± 0.44 & 1.11 ± 0.72 & \textbf{1.05 ± 0.85} & 1.43 ± 1.22 & \textbf{1.03 ± 0.93} \\ 
\midrule
\multirow{3}{*}{German} & Accuracy & 72.58 ± 0.59 & 73.69 ± 0.50 & \textbf{74.90 ± 0.20} & 72.84 ± 0.26 & 71.78 ± 0.37 & 74.04 ± 0.36 & 73.02 ± 0.46 & \textbf{74.27 ± 0.09} \\ 
& $\Delta_{DP}$ & 18.82 ± 7.48 & 4.89 ± 1.04 & 0.99 ± 0.90 & 9.73 ± 1.42 & 11.13 ± 1.40 & \textbf{0.02 ± 0.05} & 5.94 ± 2.55 & \textbf{0.12 ± 0.23} \\ 
& $\Delta_{EO}$ & 15.53 ± 5.72 & 4.58 ± 0.38 & 0.78 ± 0.73 & 7.72 ± 1.60 & 11.94 ± 1.73 & \textbf{0.05 ± 0.10} & 6.45 ± 2.67 & \textbf{0.17 ± 0.35} \\ 
\bottomrule
\end{tabular}
}
\caption{Performance Comparisons of Graph Contrastive Learning on Synthetic Graphs
using Node Classification, where Graphair is a fair graph contrastive learning method \citep{ling2023learning}. Results with the top two values of all metrics among all methods are highlighted in bold. }
\label{tab:additional_results_nc}
\end{table*}

\begin{table*}[ht]
\centering
\resizebox{\textwidth}{!}{%
\begin{tabular}{ccccccccccc}
\toprule
\textbf{Dataset} & \textbf{Metric (\%)} & \textbf{GCL} & \textbf{Graphair} & \textbf{FairAdj} & \textbf{FairDrop (0.5)} & \textbf{FairDrop (0.75)} & \textbf{FairGen} & \textbf{FairWire} & \textbf{Diffusion} & \textbf{FairGDiff} \\ 
\midrule
\multirow{4}{*}{NBA} & Micro-F1 & 54.37 ± 0.00 & 65.05 ± 0.00 & 66.02 ± 0.00 & 62.14 ± 0.00 & 58.25 ± 0.00 & \textbf{30.10 ± 0.00} & 74.76 ± 0.00 & 51.46 ± 0.00 & \textbf{50.49 ± 0.00} \\ 
& $\Delta_{DP}$ & 1.51 ± 0.01 & \textbf{0.15 ± 0.01} & \textbf{0.18 ± 0.01} & 0.21 ± 0.01 & 2.07 ± 0.01 & \textbf{0.15 ± 0.01} & 1.67 ± 0.01 & 0.45 ± 0.01 & 0.43 ± 0.00 \\ 
& $\Delta_{EO}$ & 18.79 ± 6.55 & \textbf{8.78 ± 4.71} & 14.14 ± 10.37 & 11.67 ± 8.17 & 17.02 ± 8.91 & 12.99 ± 4.28 & 14.48 ± 5.16 & 16.21 ± 6.49 & \textbf{7.11 ± 3.52} \\ 
& NDCG@10 & 27.58 ± 0.61 & 26.41 ± 0.49 & 7.33 ± 0.28 & 23.01 ± 0.44 & 21.51 ± 0.31 & 18.05 ± 0.36 & 5.25 ± 0.21 & \textbf{34.17 ± 0.39} & \textbf{41.44 ± 0.30} \\ 
\midrule
\multirow{4}{*}{German} & Micro-F1 & 89.45 ± 0.69 & 90.50 ± 0.00 & 70.00 ± 0.00 & 85.50 ± 0.00 & 79.50 ± 0.00 & \textbf{63.00 ± 0.00} & 77.00 ± 0.00 & 67.90 ± 1.71 & \textbf{66.80 ± 1.40} \\ 
& $\Delta_{DP}$ & 2.82 ± 0.64 & 1.84 ± 0.01 & \textbf{0.16 ± 0.01} & 8.34 ± 0.01 & 8.28 ± 0.02 & 1.16 ± 0.01 & 0.30 ± 0.00 & 0.39 ± 0.32 & \textbf{0.04 ± 0.08} \\ 
& $\Delta_{EO}$ & 26.28 ± 19.68 & 12.67 ± 4.59 & 9.62 ± 3.64 & 29.38 ± 7.32 & 35.83 ± 16.14 & 18.17 ± 10.90 & \textbf{6.42 ± 2.29} & 20.95 ± 23.76 & \textbf{4.95 ± 9.90} \\ 
& NDCG@10 & \textbf{28.70 ± 1.62} & 26.08 ± 0.21 & 9.84 ± 0.15 & 17.01 ± 0.23 & 15.72 ± 0.29 & 15.12 ± 0.17 & 9.04 ± 0.18 & 22.78 ± 15.75 & \textbf{31.88 ± 7.43} \\ 
\bottomrule
\end{tabular}
}
\caption{Performance Comparisons of Graph Contrastive Learning methods on Synthetic Graphs using Link Prediction. Results with the top two values of all metrics among all methods are highlighted in bold. }
\label{tab:additional_results_lp}
\end{table*}

\section{Experiments}
\label{sec:experiment}

\subsection{Experiment Settings}
\noindent{\textbf{Datasets.}}
In our experiments, we use four real-world datasets: NBA~\citep{dai2021say}, German Credit~\citep{asuncion2007uci}, and Pokec (including both Pokec-n and Pokec-z)~\citep{takac2012data}. NBA records basketball player statistics with race as the sensitive attribute, while German Credit contains financial records with gender as the sensitive attribute, commonly used for fairness evaluation in credit scoring. Pokec-n and Pokec-z are regional subsets of the Slovak social network Pokec, where gender is the sensitive attribute, making them suitable for studying biases in social link formation. These datasets provide diverse scenarios for evaluating fairness in graph generation. 

\noindent{\textbf{Graph Diffusion Models.}} For the diffusion model for training in \textit{FairGDiff}, we utilize a latent graph diffusion model~\citep{evdaimon2024neural} in our experiments due to computational resource constraints limiting the feasibility of other graph diffusion models discussed in Sec.~\ref{subsec:graph_diffusion}.

\noindent{\textbf{Baselines.}}
We mainly compare our experimental results with four fair graph generative models, namely FairAdj \citep{li2021dyadic}, FairDrop \citep{spinelli2021fairdrop}, FairWire~\citep{kose2024fairwire}, and FairGen~\citep{zheng2024fairgen}. FairAdj~\citep{li2021dyadic} aims to obtain a fair adjacency matrix under some fairness restrictions for link prediction. FairDrop~\citep{spinelli2021fairdrop} finds a debiased adjacency matrix by randomly dropping some biased edges in graph data. FairWire~\citep{kose2024fairwire} is a diffusion-based graph generation framework that embeds a theory-guided fairness regularizer into the denoiser to mitigate structural bias in both link prediction and synthetic graphs. FairGen~\citep{zheng2024fairgen} defines a supervised random walk strategy to avoid biased connections by incorporating both label information and parity constraints.

\noindent{\textbf{Environment Settings.}}
Our experiments are conducted on a system equipped with an Intel Xeon® Platinum 8255C CPU (12 cores), 40GB RAM, and an NVIDIA GeForce RTX 2080 Ti GPU (11GB VRAM). The software environment includes CUDA 12.4, PyTorch 2.0.1, and Python 3.8, ensuring efficient GPU acceleration for deep learning tasks.


\subsection{Evaluation Metrics}
\noindent - \textbf{\textit{Metrics for Utility.}}
In our experiments, we measure utility using two common metrics: test accuracy for node classification and NDCG@10~\citep{jarvelin2002cumulated} for link prediction.

\noindent - \textbf{\textit{Metrics for Fairness.}}
For fairness evaluation, we adopt statistical parity ($\Delta_{DP}$) and equal opportunity ($\Delta_{EO}$), defined as follows:
($\Delta_{DP}$) and equal opportunity ($\Delta_{EO}$), defined as below: 
\[ \Delta_{DP} = |P(\hat{y}=1|s=0)-P(\hat{y}=1|s=1)|, \]
\[ \Delta_{EO} = |P(\hat{y}=1|y=1,s=0)-P(\hat{y}=1|y=1,s=1)|, \]
where $s$ represents the binary sensitive attribute, and $\hat{y}$ is the predicted label in distinguish to the realistic label $y$. 

\noindent - \textbf{\textit{Metrics for Efficiency.}}
To evaluate the efficiency of \textit{FairGDiff}, we use Floating Point Operations Per Second (FLOPs)~\citep{zaharia2016apache} as a hardware-independent metric for computational complexity.

\subsection{Node Classification on Synthetic Graphs}\label{subsec:node_classification}
To assess the efficacy of \textit{FairGDiff}, we conduct node classification experiments on generated graphs  generated by \textit{FairGDiff} and baseline methods. We evaluate on NBA~\citep{dai2021say}, Pokec-n, and Pokec-z~\citep{takac2012data} datasets. For each dataset, nodes are randomly split into 50\% for training, 25\% for validation, and 25\% for testing. A GCN classifier is trained on the generated graphs  to predict node labels, enabling a fair comparison across generative models. 

The results shown in Table~\ref{tab:additional_results} demonstrate that
\textit{FairGDiff} consistently outperforms baseline methods with better trade-off between utility (Accuracy) and fairness ($\Delta_{DP}$ and $\Delta_{EO}$) across three different datasets. This implies that our approach mitigates topology-induced bias while preserving the predictive performance of synthetic graphs. 

\subsection{Graph Contrastive Learning on Synthetic Graphs}\label{subsec:contrastive_learning}
Since \textit{FairGDiff} is trained without node label information, it naturally aligns with self-supervised learning frameworks, making Graph Contrastive Learning (GCL)~\citep{you2020graph} a suitable evaluation approach. GCL learns node representations by contrasting similar (positive) and dissimilar (negative) node pairs, allowing us to assess whether \textit{FairGDiff} generates embeddings that retain structural utility while mitigating bias.

To be specific, we first learn node representations from all generated graphs  using graph contrastive learning. To evaluate the quality of the learned representations, we conduct two downstream tasks: node classification and link prediction. For node classification, we train a GCN classifier using embeddings obtained from graph contrastive learning. In terms of link prediction, we use the learned embeddings to predict missing links in the graph.
Here, we conduct experiments on NBA~\citep{dai2021say}, Pokec-n, Pokec-n~\citep{takac2012data}, and German Credit~\citep{asuncion2007uci} datasets. For Pokec-n and Pokec-z, we randomly split 10\%, 10\%, and 80\% for training, validating, and testing, and the mini-batch training pipeline is used to reduce computation complexity for these large scaled datasets. For NBA  \citep{dai2021say} and German Credit~\citep{asuncion2007uci}, a 20\%, 35\%, and 45\% splitting ratio is allocated to training, validating, and testing, and a whole-batch training strategy is adopted with them. 

The trade-off performance for node classification and link prediction by GCL is presented in Table~\ref{tab:additional_results_nc} and Table~\ref{tab:additional_results_lp} respectively. For node classification, the evaluation is conducted on embeddings learned via GCL rather than directly using raw node features. This setup allows us to assess whether the learned representations from generated graphs  effectively capture structural information while mitigating bias. We also evaluate the effectiveness of embeddings learned through GCL for link prediction tasks by measuring how well they preserve meaningful connectivity patterns. The results in both cases indicate not only a superior utility–fairness trade-off over baselines, but also some gains in individual utility and individual fairness metrics, particularly on the link prediction task.

\subsection{Efficiency Analysis}
To evaluate the efficiency of the proposed \textit{FairGDiff}, we compare \textit{FairGDiff} with one-step baselines across four different datasets. Results shown in Table~\ref{tab:efficiency} demonstrate that \textit{FairGDiff} significantly reduces computational cost compared to baselines, highlighting its efficiency in fair graph generation. \textit{FairGDiff} maintains low cost across all datasets, demonstrating both its efficiency and scalability.


\begin{table}[h!]
    \centering
    \caption{(RQ3) Computational cost comparison by FLOPs on different Datasets.}
    \label{tab:efficiency}
    \resizebox{0.59\columnwidth}{!}{%
    \begin{tabular}{ccccc}
        \toprule
        \textbf{FLOPs ($\times 10^9$)} & \textbf{NBA} & \textbf{German Credit} & \textbf{Pokec-n} & \textbf{Pokec-z} \\
        \midrule
        \textbf{FairGDiff} & \textbf{9.0} & \textbf{19.3} & \textbf{172.5} & \textbf{181.7} \\
        \textbf{FairGen} & $36.1$ & $113.2$ & $211.2$ & $299.8$ \\
        \textbf{FairWire} & $1,820.0$ & $3,939.0$ & $12,580.0$ & $800,800.0$\\
        \bottomrule
    \end{tabular}
    }
\end{table}

\subsection{Topology Bias Ratios}
To assess the performance of \textit{FairGDiff}, we report the topology-bias ratios $\mathcal{T}_0(G)$ and $\mathcal{T}_1(G)$ for both the original biased graph and the \textit{FairGDiff} synthetic graph in Table~\ref{tab:tb_stats}; as shown, \textit{FairGDiff} consistently drives both ratios toward the independence target $\mathcal{T}_x(\hat{G})=1$ ($x\in\{0,1\}$), indicating effective mitigation of targeted topology biases. 

\begin{table}[h!]
    \centering
    \caption{Topology biases compared between the original biased graph and the \textit{FairGDiff} synthetic graph.}
    \label{table:efficiency}

    \resizebox{0.59\columnwidth}{!}{%
    \begin{tabular}{ccccc}
        \toprule
        $\mathcal{T}_0(G)$ & \textbf{NBA} & \textbf{German Credit} & \textbf{Pokec-n} & \textbf{Pokec-z} \\
        \midrule
        \textbf{Original Graph} & $0.930$ & $0.958$ & \textbf{1.000} & \textbf{1.000} \\
        \textbf{FairGDiff Synthetic Graph} & \textbf{0.987} & \textbf{0.982} & \textbf{1.000} & \textbf{1.000} \\
        \bottomrule
    \end{tabular}
    }
    \resizebox{0.59\columnwidth}{!}{%
    \begin{tabular}{ccccc}
        \toprule
        $\mathcal{T}_1(G)$ & \textbf{NBA} & \textbf{German Credit} & \textbf{Pokec-n} & \textbf{Pokec-z} \\
        \midrule
        \textbf{Original Graph} & $1.404$ & $3.088$ & $15.439$ & $17.748$ \\
        \textbf{FairGDiff Synthetic Graph} & \textbf{1.039} & \textbf{1.186} & \textbf{11.040} & \textbf{13.481} \\
        \bottomrule
    \end{tabular}
    }
    \label{tab:tb_stats}
\end{table}

\section{Conclusion}

In this work, we propose \textit{FairGDiff}, a novel fairness-aware graph generation framework that mitigates topology biases in graph diffusion through counterfactual intervention. 
It provides a one-step, computationally efficient solution and extensive experiments on multiple real-world datasets show that our method outperforms prior approaches in achieving an optimal trade-off between fairness and downstream task performance while maintaining scalability.

\bibliographystyle{ACM-Reference-Format}
\bibliography{sample-base}

\appendix

\section{Detailed Datasets Information}
In this section, we provide detailed information about the datasets we use in the experiment section. All the datasets are publicly available social networks, with statistical details provided in Table \ref{tab:dataset_statistics}.
\begin{table}[h!]
    \centering
    \begin{tabular}{lcccc}
        \toprule
        \textbf{Statistics} & \textbf{NBA} & \textbf{German} & \textbf{Pokec-n} & \textbf{Pokec-z} \\
        \midrule
        \# of nodes & 403 & 1000 & 66,569 & 67,796 \\
        \# of edges & 16,570 & 22,242 & 729,129 & 882,765 \\
        Density & 0.10203 & 0.02224 & 0.00016 & 0.00019 \\
        \bottomrule
    \end{tabular}
    \caption{Statistics for Different Graph Datasets}
    \label{tab:dataset_statistics}
\end{table}

\section{Graph Property Analysis}

\begin{figure*}[h!]
    \centering
    \includegraphics[width=0.99\textwidth]{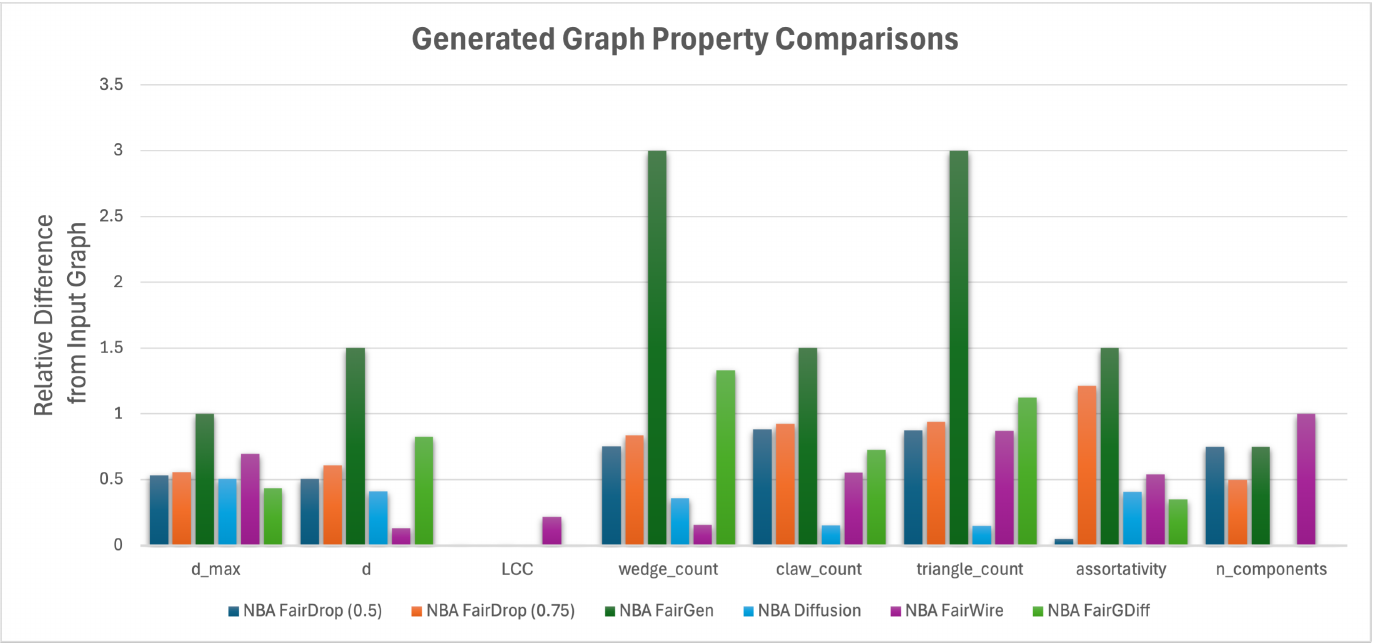} 
    \caption{Graph Property Comparisons on NBA Dataset \citep{dai2021say}. Relative differences from input graph are calculated as follows as: $\frac{|M(G^F)-M(\hat{G})|}{|M(G^F)|}$, where $G^F$ is the input graph, $\hat{G}$ is the generated graph, and $M$ stands for a certain metric shown in Table \ref{tab:metric_computations}. }
    \label{fig:subfig9}
\end{figure*}

The following are some graph property analyses on all generated graphs. The meanings and representations of the graph properties are listed in Table \ref{tab:metric_computations}. From Figure \ref{fig:subfig9} we can conclude that compared to other fair graph generative models and diffusion models, our FairGDiff preserves most graph properties of the input graph when applying debiasing methods in graph diffusion. 


\begin{table}[h!]
\centering
\resizebox{0.9\textwidth}{!}{%
\begin{tabular}{cc}
\toprule
\textbf{Metric Name} & \textbf{Computation} \\ 
\midrule
Max Degree & $\max \deg(v)$ where $v \in \mathcal{V}$ \\ 
\hline
Average Degree & $\frac{2|\mathcal{E}|}{|\mathcal{V}|}$ \\ 
\midrule
Largest Connected Component & $\max_{C \in \mathcal{C}} |C|$ \\ 
\midrule
Wedge Count & $\sum_{v \in \mathcal{V}} \binom{\deg(v)}{2}$ \\ 
\midrule
Claw Count & $\sum_{v \in \mathcal{V}} \binom{\deg(v)}{3}$ \\ 
\midrule
Triangle Count & $\frac{1}{3} \sum_{v \in \mathcal{V}} T(v)$, where $T(v)$ is the number of triangles incident to $v$ \\ 
\midrule
Gini Coefficient & $1 - 2 \int_0^1 L(x) dx$, where $L(x)$ is the Lorenz curve \\ 
\midrule
Relative Edge Distribution Entropy & $-\sum_{v \in \mathcal{V}} \frac{\deg(v)}{2|\mathcal{E}|} \log \left( \frac{\deg(v)}{2|\mathcal{E}|} \right)$ \\ 
\midrule
Clustering Coefficient & $\frac{\sum_{v \in \mathcal{V}} T(v)}{\sum_{v \in \mathcal{V}} \binom{\deg(v)}{3}}$, where $T(v)$ is the number of triangles incident to $v$ \\
\midrule
Assortativity Coefficient & $\frac{\sum_{xy} A_{xy} (x - \bar{x})(y - \bar{y})}{\sigma_x \sigma_y}$ \\ 
\midrule
\# of Connected Components & $\text{Number of connected components in the graph}$ \\ 
\bottomrule
\end{tabular}
}
\caption{Metric Computations for Graph Properties}
\label{tab:metric_computations}
\end{table}

\section{Distribution Analysis of Sensitive Attributes}
\begin{figure*}[ht]
    \centering
    
    \parbox[b]{0.45\textwidth}{
        \centering
        \includegraphics[width=\linewidth]{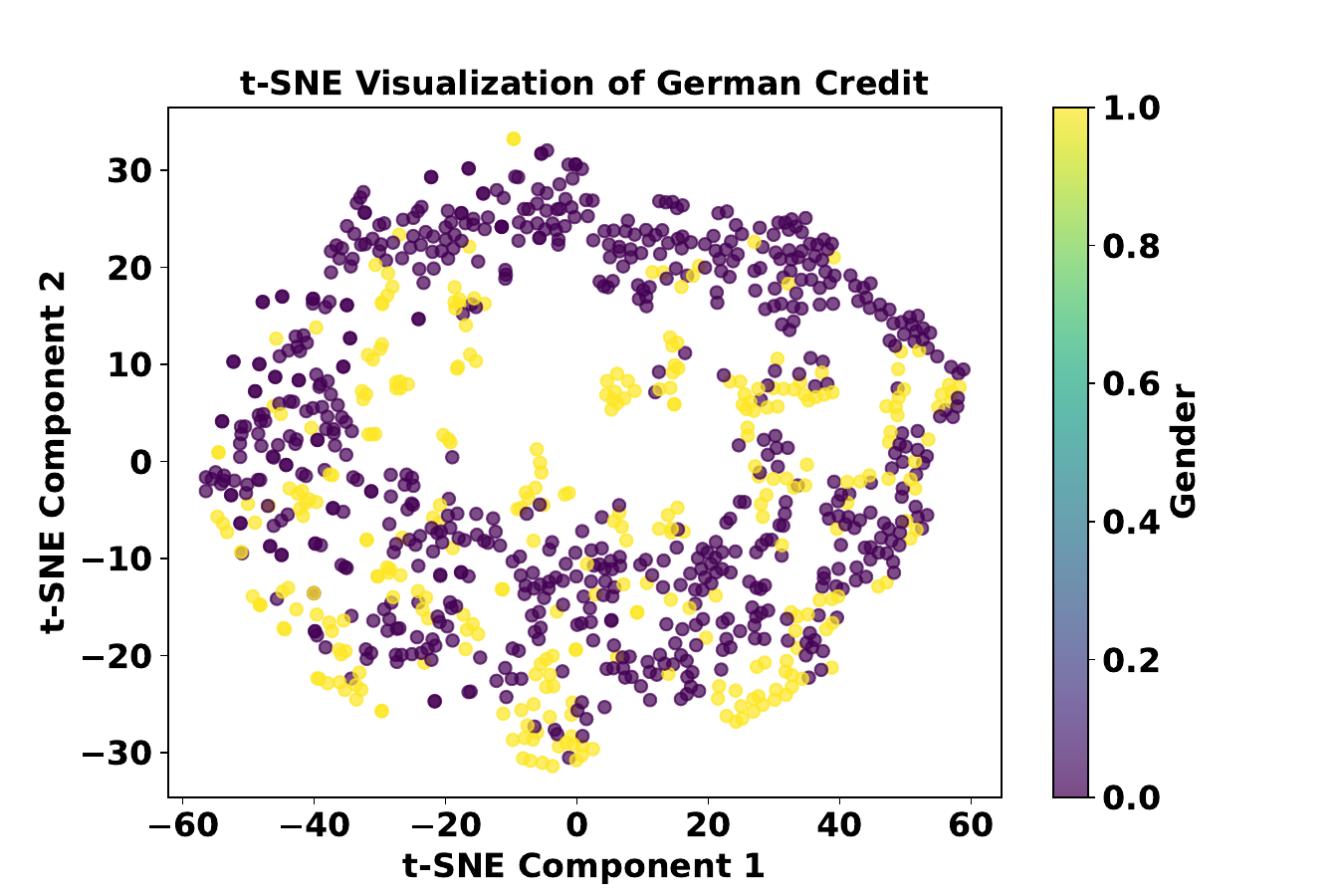}
        \captionof{figure}{Graph embedding of input biased graph.}
        \label{fig:embedding_original}
    }
    \hfill
    \parbox[b]{0.45\textwidth}{
        \centering
        \includegraphics[width=\linewidth]{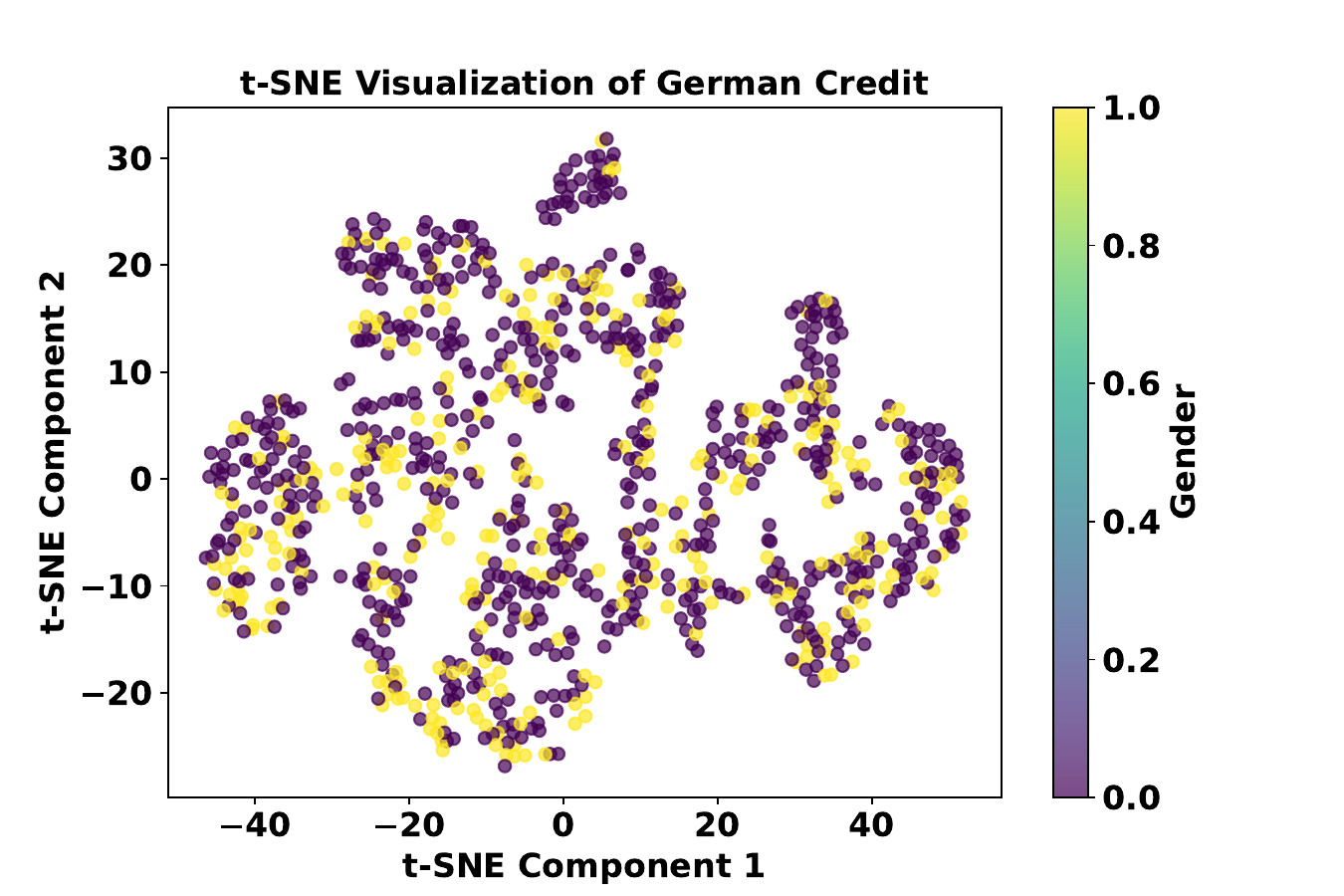}
        \captionof{figure}{Graph embedding by FairGDiff.}
        \label{fig:embedding_FairGDiff}
    }
    
    \vspace{0.5cm} 
    
    \parbox[b]{0.45\textwidth}{
        \centering
        \includegraphics[width=\linewidth]{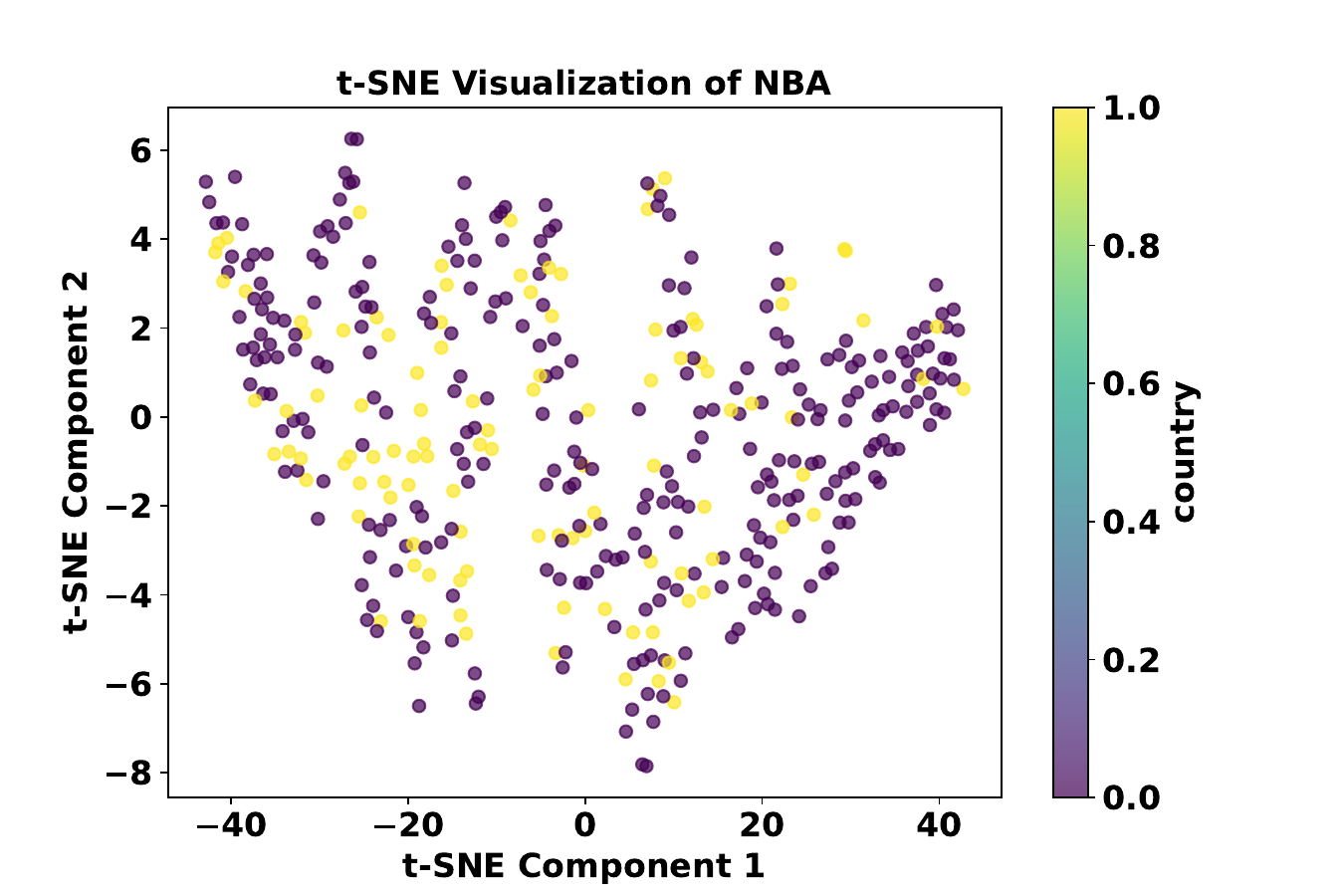}
        \captionof{figure}{Graph embedding of input biased graph (NBA).}
        \label{fig:embedding_original_NBA}
    }
    \hfill
    \parbox[b]{0.45\textwidth}{
        \centering
        \includegraphics[width=\linewidth]{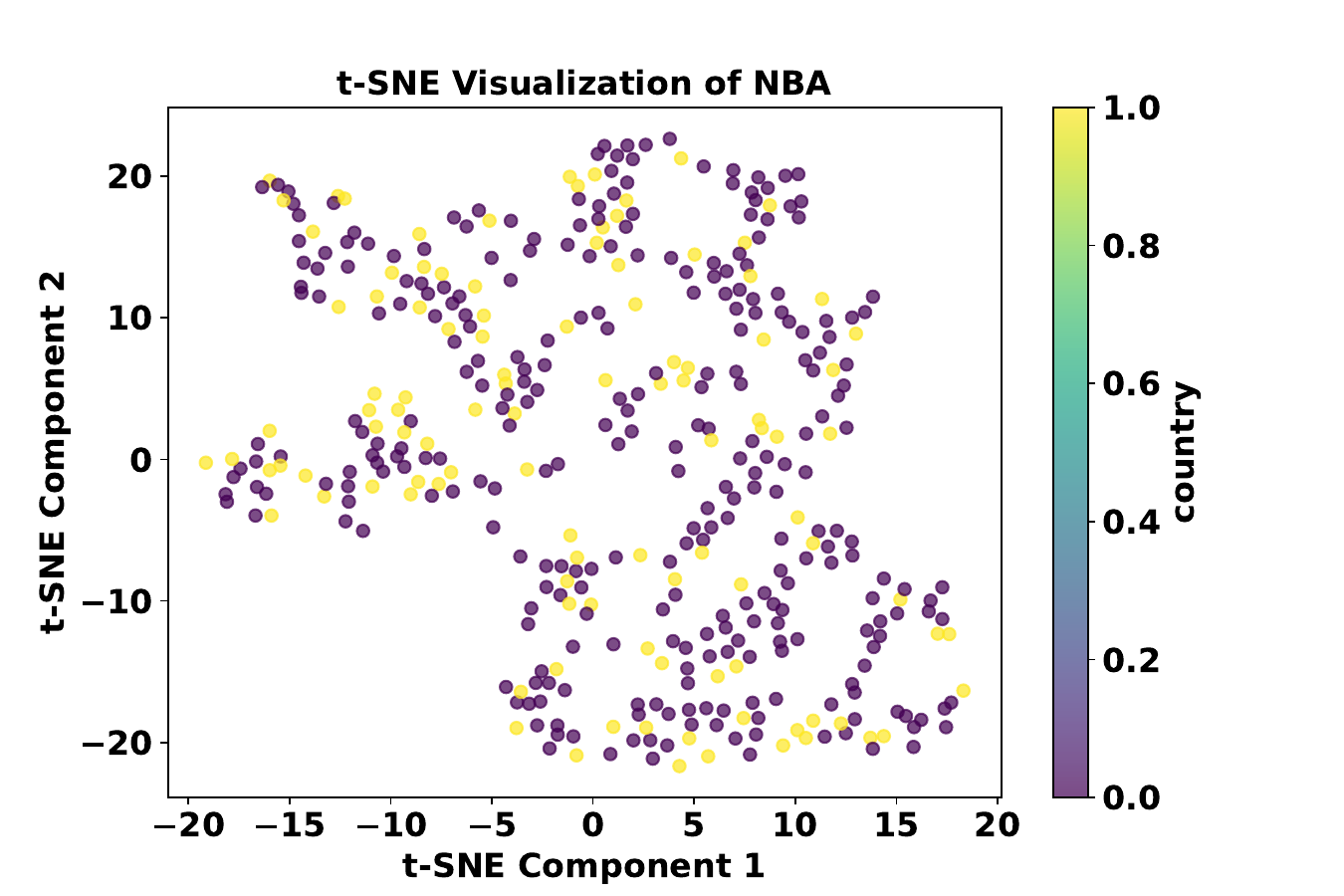}
        \captionof{figure}{Graph embedding by FairGDiff (NBA).}
        \label{fig:embedding_FairGDiff_NBA}
    }
    
    \caption{Overall comparison between input biased graphs and FairGDiff-generated graphs.}
    \label{fig:embedding_comparison}
    
\end{figure*}

We leverage the quantitative analysis in Table \ref{tab:quant} and the visualization in Figure \ref{fig:embedding_comparison} to examine the distribution gap between nodes with different sensitive attribute values in both the input biased graph and the generated graph of our proposed \textit{FairGDiff}.

We use the Wasserstein distance to quantify the distribution gap between nodes in both graphs, where each node is represented by a 64-dimensional vector in the node embedding space. The node embeddings are obtained through graph contrastive learning using non-sensitive node attributes and the adjacency matrix. A smaller Wasserstein distance between nodes in our generated graph compared to the input biased graph indicates higher fairness, demonstrating the efficacy of \textit{FairGDiff}. 

\begin{table}[h!]
    \centering
    \caption{Quantitative analysis of sensitive attributes.}
    \resizebox{0.6\columnwidth}{!}{%
        \begin{tabular}{ccccc}
            \toprule
            \textbf{The Wasserstein Distance} & \textbf{NBA} & \textbf{German Credit} \\
            \midrule
            \textbf{Input Biased Graph} & $0.838$ & $2.066$ \\
            \textbf{\textit{FairGDiff} Generated Graph} & $0.209$ & $0.854$ \\
            \bottomrule
        \end{tabular}
    }
    \label{tab:quant}
\end{table}

We also demonstrate the efficacy of our proposed method \textit{FairGDiff} more intuitively, where we use t-SNE~\citep{van2008visualizing} to visualize the node embedding trained from graph contrastive learning on both graphs using NBA and German Credit datasets. The node embeddings are projected to 2-D space as shown in Figure~\ref{fig:embedding_comparison}. Two colors are used in Figure~\ref{fig:embedding_comparison} to represent binary sensitive attribute values. For NBA dataset, yellow dots represent \textit{from the US} and purple dots representing \textit{not from the US}. For German Credit dataset, yellow dots stand for \textit{female}, while purple dots stand for \textit{male}. 

We provide the reference node embeddings in Figure~\ref{fig:embedding_original} and Figure~\ref{fig:embedding_original_NBA} trained from graph contrastive learning on the original German Credit and NBA graphs respectively. The results in Figure~\ref{fig:embedding_original} and Figure~\ref{fig:embedding_original_NBA} show that the distributions of the two sensitive attribute groups in the original graph exhibit a significant shift, indicating a strong separation in node embeddings based on the gender attribute. In contrast, the distribution gap between the two sensitive attribute groups in the synthetic graph by our method as shown in Figure~\ref{fig:embedding_FairGDiff} and Figure~\ref{fig:embedding_FairGDiff_NBA} is significantly reduced, with node embeddings appearing more blended and less segregated by corresponding sensitive attribute.
This demonstrates that \textit{FairGDiff} effectively reduces the distribution shift of sensitive attributes, mitigating bias and promoting fairness in the synthetic graph.

\section{Additional Experiment Results}
We additionally provide some utility-fairness trade-off results demonstrated by the hypervolume metric. The Hypervolume metric~\citep{zitzler1999multiobjective} is denoted as $\mathcal{H(\cdot)}$, which is computed based on the combination of utility and fairness metrics. We use $\mathcal{H} (\text{Acc}, \Delta_{DP}, \Delta_{EO})$ for node classification and $\mathcal{H}(\text{NDCG@10}, \Delta_{DP}, \Delta_{EO})$ for link prediction. Specifically, we have the following hypervolume calculation: $\mathcal{H} (\text{Acc}, \Delta_{DP}, \Delta_{EO})=|\text{Acc}-\text{Acc}_{\text{ref}}| \cdot |\Delta_{DP}-\Delta_{DP, \text{ref}}| \cdot |\Delta_{EO}-\Delta_{EO, \text{ref}}|$ and $\mathcal{H} (\text{NDCG@10}, \Delta_{DP}, \Delta_{EO})=|\text{NDCG@10}-\text{NDCG@10}_{\text{ref}}| \cdot |\Delta_{DP}-\Delta_{DP, \text{ref}}| \cdot |\Delta_{EO}-\Delta_{EO, \text{ref}}|$. We set each $*_{\text{ref}}$ to the worst (least favorable) value of its corresponding metric across all baselines and FairGDiff. 

The results shown in Table~\ref{tab:node_classification} demonstrate that
\textit{FairGDiff} consistently outperforms baseline methods with higher $\mathcal{H} (\text{Acc}, \Delta_{DP}, \Delta_{EO})$ across three different datasets. This implies that our approach obtains better trade-off performance between utility and fairness, mitigating topology-induced bias while preserving the predictive performance of synthetic graphs. \begin{table*}[ht]
\centering
\caption{Trade-off performance for the node classification task. *FairWire~\citep{kose2024fairwire} uses smaller-scaled Pokec-n and Pokec-z datasets compared to our FairGDiff.}
\label{tab:node_classification}
\resizebox{0.95\textwidth}{!}{%
\begin{tabular}{cccccccccc}
\toprule
\textbf{Dataset} & \textbf{MLP} & \textbf{GCN} & \textbf{FairAdj} & \textbf{FairDrop (0.5)} & \textbf{FairDrop (0.75)} & \textbf{FairGen} & \textbf{FairWire} & \textbf{Diffusion} & \textbf{FairGDiff} \\
\midrule
\textbf{NBA}   &    1.2434 & 0.0086 & 0.0000 & 2.6625 & 2.7556 & 2.4764 & 2.3242 & 0.0760 & \textbf{3.8808} \\
\textbf{Pokec-n}   & 3.8249 & 4.5700 & 4.6341 & 3.6447 & 2.4842 & 4.4615 & $-^{\ast}$ & 5.1858 & \textbf{5.4182} \\
\textbf{Pokec-z}   & 3.9150 & 4.0534 & 4.5960 & 2.7388 & 0.0000 & 5.6833 & $-^{\ast}$     & 4.7156 & \textbf{5.8601} \\
\bottomrule
\end{tabular}}
\vspace{-3mm}
\end{table*}

The trade-off performance for both node classification and link prediction by GCL is presented in Table~\ref{tab:contrastive_learning}.
For node classification, the evaluation is conducted on embeddings learned via GCL rather than directly using raw node features. This setup allows us to assess whether the learned representations from generated graphs  effectively capture structural information while mitigating bias. A higher hypervolume score $\mathcal{H} (\text{Acc}, \Delta_{DP}, \Delta_{EO})$ indicates that \textit{FairGDiff} not only maintains classification utility but also reduces fairness disparities in the embedding space.

\begin{table*}[h]
    \centering
    \caption{ Trade-off performance for node classification and link prediction with graph contrastive learning in Sec.~\ref{subsec:contrastive_learning}. }
    \label{tab:contrastive_learning}
    \resizebox{0.95\textwidth}{!}{%
    \begin{tabular}{cccccccccc}
        \toprule
        \textbf{Dataset} &  \textbf{GCL} & \textbf{Graphair} & \textbf{FairAdj} & \textbf{FairDrop (0.5)} & \textbf{FairDrop (0.75)} & \textbf{FairGen} & \textbf{Diffusion} & \textbf{FairGDiff} \\
        \midrule
        \multicolumn{9}{c}{\textbf{Node Classification} ($\mathcal{H} (\text{Acc}, \Delta_{DP}, \Delta_{EO})$)}
        \\ \midrule
        \textbf{NBA} 
         & 2.9669 & 0.0002 & 1.0963 & 5.6109 & 7.5334 & 3.6595 & 0.0061 & \textbf{11.6733} \\
        \textbf{Pokec-z} 
         & 0.2975 & 0.0003 & 0.5304 & 0.2942 & 0.4012 & 0.3247 & 0.3089 & \textbf{0.5134} \\
        \textbf{German} 
         & 0.0002 & 4.5838 & 8.2017 & 2.0840 & 0.7900 & 8.8193 & 3.4413 & \textbf{8.7715} \\
         \midrule
        \multicolumn{9}{c}{\textbf{Link Prediction} ($\mathcal{H}(\text{NDCG@10}, \Delta_{DP}, \Delta_{EO})$)}
        \\ \midrule
         \textbf{NBA} 
         & 0.0003 & \textbf{1.1264} & 0.5245 & 0.7377 & 0.0027 & 0.2572 & 0.1883 & \textbf{0.8412} \\
        \midrule
        \textbf{German} 
         & 8.7425 & \textbf{23.1210} & 22.2711 & 2.8664 & 0.0686 & 12.3626 & 12.0645 & \textbf{25.0539} \\
        \bottomrule
    \end{tabular}}
\end{table*} 

Similarly, we also evaluate the effectiveness of embeddings learned through GCL for link prediction tasks by measuring how well they preserve meaningful connectivity patterns. The results present that \textit{FairGDiff} achieves very high hypervolume scores $\mathcal{H}(\text{NDCG@10}, \\ \Delta_{DP}, \Delta_{EO})$ compared to baselines, indicating its effectiveness in producing unbiased yet informative representations for link prediction. 


\section{Parameter Sensitivity Analysis}
There are two essential hyperparameters in FairGDiff, namely $\gamma_1$ controlling the utility of synthetic graphs and $\gamma_2$ alleviating the bias during the denoising process. We train FairGDiff on NBA with different $\gamma_1$ and $\gamma_2$ to determine the parameter sensitivity. With the sensitivity analysis, we conclude the best ranges to achieve a high utility score with good fairness constraints. 

\begin{figure*}[ht]
    \centering
    \small
    \begin{center}
    \includegraphics[width=0.99\textwidth]{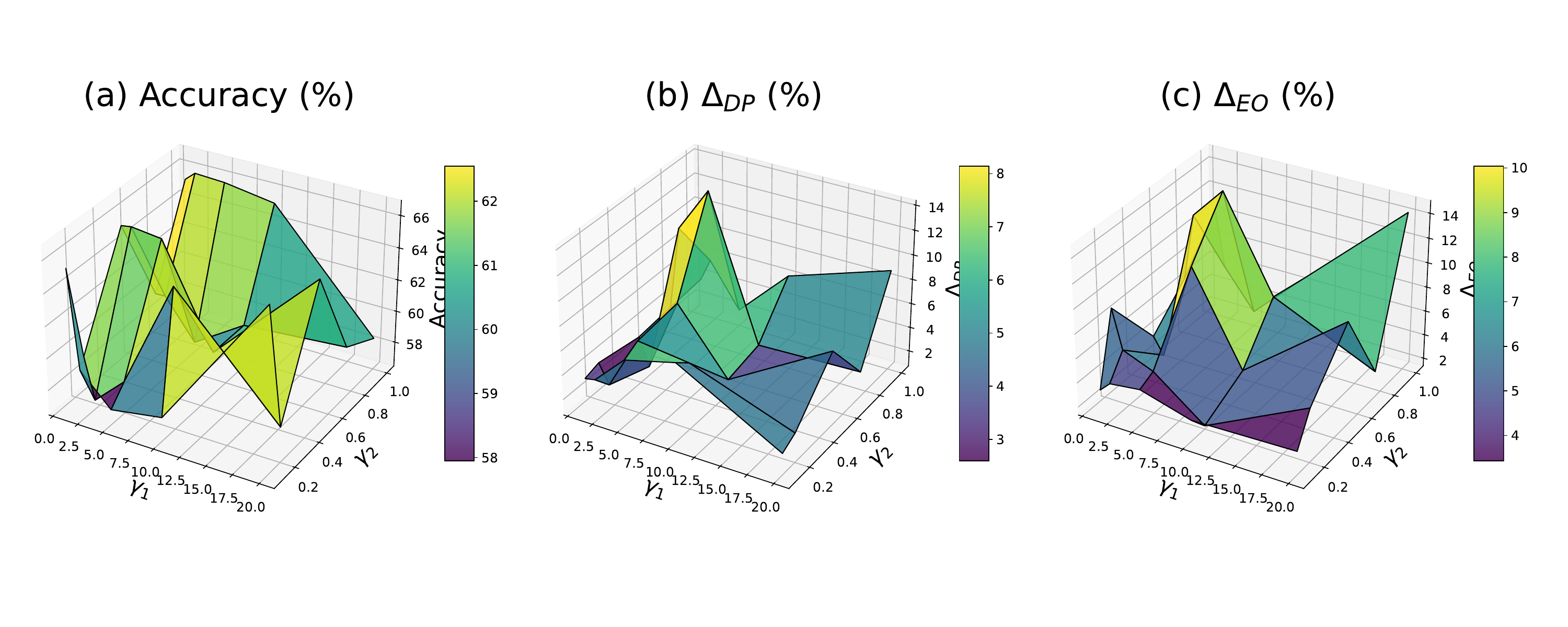} 
    \end{center}
    \vspace{-5em}
    \caption{Parameter sensitivity analysis of FairGDiff on NBA.}
    \label{fig:sens}
    \normalsize
\end{figure*}

We set the hyperparameters $\gamma_1$ in $[1,2,5,10,20]$ and $\gamma_2$ in $[0.1,0.2,\\0.5,0.75,1]$ via node classification on NBA. We can find that when $\gamma_1 \leq 5$ and $\gamma_2 \leq 0.5$, FairGDiff reaches the best balance between classification accuracy and fairness metrics. The detailed results of the sensitivity analysis are shown in Figure \ref{fig:sens}, where we conclude the best choice of $\gamma_1$ and $\gamma_2$ to be $\gamma_1=5$ and $\gamma_2=0.2$.  

\section{Limitations}
Although our experiments follow previous works and utilize a latent graph diffusion model for training due to computational constraints, this does not imply that other types of graph diffusion models cannot be easily adapted in \textit{FairGDiff}. As computational resources advance, future work can explore the integration of alternative diffusion models to further enhance performance and flexibility.

\end{document}